%% file: DrueSeif_22.tex
\documentclass[AMS,STIX1COL]{WileyNJD-v2} 

\articletype{Original Paper}%

\received{}
\revised{}
\accepted{}

\usepackage{amssymb}

\usepackage{graphicx}
\usepackage{makecell}
\usepackage{siunitx}
\usepackage{bm}
\usepackage{xspace}
\usepackage{tabularx,booktabs}

\newcommand{\uffw}{\ub_{\rm ffw}}
\newcommand{\nc}{n_{\rm c}}
\newcommand{\trans}{{\mathrm{T}}}

\graphicspath{{Figures/},{tikz/}}
\newlength\figureheight 
\newlength\figurewidth 

\usepackage{setspace}
\usepackage{tabularx}
\newcolumntype{L}[1]{>{\raggedright\arraybackslash}p{#1}} 
\newcolumntype{C}[1]{>{\centering\arraybackslash}p{#1}} 
\newcolumntype{R}[1]{>{\raggedleft\arraybackslash}p{#1}} 
\usepackage{amsmath}
\usepackage{tikz}
\usetikzlibrary{spy}
\usetikzlibrary{external}
\usepackage{pgfplots,pgfplotstable}
\usetikzlibrary{matrix}
\usepgfplotslibrary{groupplots}
\pgfplotsset{compat=newest}
\pgfplotsset{
    every x tick label/.append style = {font=\footnotesize},
    every y tick label/.append style = {font=\footnotesize},
    }
\usetikzlibrary{arrows.meta}
\usepackage[export]{adjustbox}

\newcommand{\cb}{\mbox{$\bm{c}$}\xspace} 
 
\newcommand{\eb}{\mbox{$\bm{e}$}\xspace} 
\newcommand{\fb}{\mbox{$\bm{f}$}\xspace}

\newcommand{\kb}{\mbox{$\bm{k}$}\xspace}

\newcommand{\qb}{\mbox{$\bm{q}$}\xspace} 
\newcommand{\rb}{\mbox{$\bm{r}$}\xspace}

\newcommand{\ub}{\mbox{$\bm{u}$}\xspace} 
\newcommand{\vb}{\mbox{$\bm{v}$}\xspace} 
 
\newcommand{\xb}{\mbox{$\bm{x}$}\xspace} 
\newcommand{\yb}{\mbox{$\bm{y}$}\xspace} 
\newcommand{\zb}{\mbox{$\bm{z}$}\xspace} 

\newcommand{\Bb}{\mbox{$\bm{B}$}\xspace} 
\newcommand{\Cb}{\mbox{$\bm{C}$}\xspace}

\newcommand{\Kb}{\mbox{$\bm{K}$}\xspace} 
\newcommand{\Lb}{\mbox{$\bm{L}$}\xspace} 
\newcommand{\Mb}{\mbox{$\bm{M}$}\xspace}

\newcommand{\Sb}{\mbox{$\bm{S}$}\xspace}

\newcommand{\Zb}{\mbox{$\bm{Z}$}\xspace} 

\newcommand{\etab}{\mbox{$\bm{\eta}$}\xspace}

\newcommand{\lamb}{\mbox{$\bm{\lambda}$}\xspace}

\newcommand{\rhob}{\mbox{$\bm{\rho}$}\xspace}

\newcommand{\vbp}{\mbox{$\dot{\bm{v}}$}\xspace} 
 
\newcommand{\xbp}{\mbox{$\dot{\bm{x}}$}\xspace} 
\newcommand{\ybp}{\mbox{$\dot{\bm{y}}$}\xspace} 
\newcommand{\zbp}{\mbox{$\dot{\bm{z}}$}\xspace} 
\newcommand{\etabp}{\mbox{$\dot{\bm{\eta}}$}\xspace} 
\newcommand{\alp}{\mbox{${\alpha}$}\xspace}

\usepackage[caption=false,subrefformat=parens,labelformat=parens]{subfig}

\begin{document}
\input{colors}

\title{Application of Stable Inversion to Flexible Manipulators Modeled by the ANCF}

\author[1]{Svenja Drücker*}

\author[1]{Robert Seifried}

\authormark{DRÜCKER \textsc{et al}}

\address[]{\orgdiv{Institute of Mechanics and Ocean Engineering}, \orgname{Hamburg University of Technology}, \orgaddress{\state{Hamburg}, \country{Germany}}}

\corres{*Svenja Drücker, Institute of Mechanics and Ocean Engineering, Hamburg University of Technology, Ei\ss endorfer Straße 42,
21073 Hamburg, Germany \\ \email{svenja.druecker@tuhh.de}}

\abstract[Summary]{Compared to conventional robots, flexible manipulators offer many advantages, such as faster end-effector velocities and less energy consumption. However, their flexible structure can lead to undesired oscillations. Therefore, the applied control strategy should account for these elasticities. A feedforward controller based on an inverse model of the system is an efficient way to improve the performance. However, unstable internal dynamics arise for many common flexible robots and stable inversion must be applied. In this contribution, an approximation of the original stable inversion approach is proposed. The approximation simplifies the problem setup, since the internal dynamics do not need to be derived explicitly for the definition of the boundary conditions. From a practical point of view, this makes the method applicable to more complex systems with many unactuated degrees of freedom. Flexible manipulators modeled by the absolute nodal coordinate formulation (ANCF) are considered as an application example. }


\jnlcitation{\cname{%
\author{S. Drücker} and
\author{R. Seifried},} (\cyear{XXX}),
\ctitle{Application of Stable Inversion to Flexible Manipulators Modeled by the ANCF}, \cjournal{GAMM-Mitteilungen}, \cvol{XXX}.}

\maketitle

\section{Introduction}\label{sec1}

The general trend to more efficient machines often results in the design of mechatronic systems with lightweight components. However, undesired oscillations can occur due to reduced stiffness of these flexible structures. Such vibrations cannot be damped directly by the available actuators, since the actuators are usually placed at the robots joints and not on the structure itself. Therefore, the systems are underactuated with more degrees of freedom than independent control inputs. Advanced control strategies are necessary in order to prevent and reduce such oscillations with the available actuators. 

Control of flexible robots is still a challenging task due to several reasons. The behavior is highly nonlinear due to large rigid body motion and possibly even nonlinear material properties. A two degree of freedom control structure with feedforward and feedback part is a typical control strategy for such systems, see e.g. \cite{MorlockEtAl22}. The feedforward part is responsible for trajectory tracking, while the feedback part is responsible for disturbance rejection. Often, the feedforward controller is an inverse model of the system. Obtaining inverse models for complex underactuated multibody systems is not straightforward, since classical methods, such as the Byrnes-Isidori normal form \cite{Sastry99} are burdensome or impossible to derive. Moreover, the internal dynamics of common flexible multibody systems are often unstable and must be taken care of. One approach redefines the system output to yield stable internal dynamics for the new output \cite{MorlockEtAl16,RaoufEtAl13}. It is also proposed to change system parameters such as mass or moment of inertia to yield stable dynamics \cite{MorlockEtAl21,Seifried12}. Alternatively, stable inversion is proposed in \cite{ChenPaden96,DevasiaEtAl96} to obtain a bounded solution to the inverse model problem for the original non-minimum phase system. The stable inversion involves solving a  two-point boundary value problem (BVP) for the internal dynamics, which need to be derived explicitly as ordinary differential equations (ODEs) from the equations of motion. The imposed boundaries are based on the stable and unstable manifolds of the zero dynamics. The obtained system input is non-causal, in the sense that the system input starts to manipulate the system before the actual start of the desired trajectory. In order to avoid the pre- and postactuation phase, a modification of the boundary value problem is proposed in \cite{GraichenEtAl05}. 
In \cite{BruelsEtAl13} is shown that the stable inversion is directly possible for an inverse model described by the servo-constraints framework. The servo-constraints framework is introduced in \cite{BlajerKolodziejczyk04} for underactuated multibody systems yielding a set of differential-algebraic equations (DAEs). In order to avoid deriving and imposing boundary conditions altogether, it was shown in \cite{BastosEtAl13} that a similar solution is obtained by solving an optimal control problem. This optimal control problem can be posed either in terms of the explicit internal dynamics in ODE form or in the servo-constraints framework. A first comparison of the boundary value problem approach with the optimal control problem is performed in \cite{BastosEtAl17}, where it is proven that the solution of the optimal control problem converges to the solution of the boundary value problem as pre- and postactuation time goes to infinity. The methodology is applied to several flexible multibody systems in \cite{LismondeEtAl19}. 

Most of the systems considered in the context of stable inversion have few unactuated degrees of freedom. The manipulator considered in \cite{BastosEtAl17} has one passive joint, while the manipulators in \cite{BastosEtAl13,DeLucaEtAl98,Seifried12} have two passive joints. For extremely light and flexible systems, models with few unactuated degrees of freedom might not be sufficient. Then, more involved mechanical models, such as finite element approaches, are necessary to accurately reflect the large deformations. 

Classical nonlinear finite elements can describe large nonlinear deformations accurately. However, typical elements cannot exactly reproduce large rigid body rotations \cite{Shabana13}. 
The floating frame of reference approach describes large nonlinear motion of a body-related reference frame. Small linear-elastic deformations are considered with respect to the reference frame \cite{Schwertassek99}. In the context of stable inversion, elastic manipulators are considered in~\cite{BurkhardtEtAl15} and~\cite{Seifried14} with six and 18 unactuated elastic degrees of freedom, respectively. However, it is not possible with the floating frame of reference approach to efficiently consider large nonlinear deformations. In contrast, the absolute nodal coordinate formulation (ANCF) is proposed in \cite{Shabana98} to accurately model large nonlinear deformations and rotations. The ANCF is applied to describe the motion of cables, flexible pendulums \cite{EscalonaEtAl98}, rubber chains \cite{MaquedaEtAl10} and tyres~\cite{SugiyamaSuda09}. It has also been applied to model flexible manipulators \cite{TianEtAl07,VoharEtAl08}. However, due to the model complexity and the difficulties in explicitly deriving its internal dynamics, ANCF beams have not yet been considered in the context of stable inversion. 

An alternative approach is taken in \cite{StroehleBetsch22} to solve the inverse dynamics problem based on servo-constraints for geometrically exact strings. Thereby, the numerical solution is based on a simultaneous space-time discretization, which avoids numerical difficulties, such as an increasing differentiation index of the underlying DAEs.

In this work, the stable inversion problem is considered for highly flexible manipulators. Thereby, the inverse model is described in the servo-constraints framework \cite{BlajerKolodziejczyk04}. An approximation of the original stable inversion problem is proposed. This approximation makes it possible to apply stable inversion to more complex systems than considered so far because an explicit derivation of the internal dynamics is avoided. It is demonstrated for a simple system with one passive joint that the solution with the proposed approximation converges to the solution of the original problem formulation. Afterwards, the methodology is applied to a flexible manipulator modeled by the ANCF for which the original formulation is not applicable.

The structure of the paper is as follows. The underlying multibody model and its inverse dynamics are described in Section~\ref{sec:2}. Stable inversion and the proposed approximation are introduced in Section~\ref{sec:3}. A convergence result supports the proposed approximation. Simulation results for a flexible manipulator modeled by the ANCF are shown in Section~\ref{sec:4}. A summary and conclusion is given in Section~\ref{sec:5}.

\section{Forward and Inverse Model}\label{sec:2}

The underlying multibody model and the framework of servo-constraints for the computation of the inverse model are introduced in the following. Holonomic underactuated systems are considered with more degrees of freedom than independent control inputs, since they naturally arise for flexible manipulators.

\subsection{Multibody Dynamics}

There exist different formulations to derive the equations of motion of general multibody systems, see e.g.~\cite{Schiehlen14}. The application of servo-constraints is independent of the specific modeling approach. Here, systems with $f$ degrees of freedom, and possibly $n_c$ geometric constraints, e.g. arising from joints or kinematic loops are considered. In a very general form, the equations of motion can be written as
\begin{align}
\ybp &= \Zb(\yb) \vb \label{eqn:ode_eqm1} \\
\Mb(\yb,t) \, \vbp + \kb(\yb,\vb,t) &= \qb(\yb,\vb,t) + \Cb(\yb,\vb,t)^{\rm T} \lamb + \Bb(\yb) \, \ub  \label{eqn:ode_eqm2} \\
\cb(\yb,\vb,t) &= \bm{0} \,.  \label{eqn:ode_eqm3}
\end{align}
Thereby, $\yb\in\mathbb{R}^{n}$ are either redundant or generalized coordinates. The matrix $\Zb\in\mathbb{R}^{n\times n}$ describes the kinematic relationship between the generalized positions $\yb$ and velocities $\vb\in\mathbb{R}^{n}$, \mbox{$\Mb\in\mathbb{R}^{n\times n}$} denotes the mass matrix,  {$\kb\in\mathbb{R}^{n}$} denotes the Coriolis and centrifugal forces, {$\qb\in\mathbb{R}^{n}$} describes the applied forces acting on the system and $\Bb\in\mathbb{R}^{n\times m}$ distributes the control input~$\ub\in\mathbb{R}^{m}$ \cite{Schiehlen14}. Equation (\ref{eqn:ode_eqm3}) describes implicit constraints $\cb\in\mathbb{R}^{\nc}$. Therefore, the system has $f=n-\nc$ degrees of freedom. In the case of redundant coordinates, these constraints arise from the joints. In the case of generalized coordinates, the constraints usually only occur if systems with kinematic loops are considered. The constraints are enforced by the Lagrange multipliers~$\lamb\in\mathbb{R}^{n_c}$ which are distributed by the Jacobian~$\Cb\in\mathbb{R}^{\nc\times \, n}$ of the constraints~$\cb$. The system output~$\zb\in\mathbb{R}^{m}$ is often chosen as the end-effector position and is defined as
\begin{align} 
\zb = \bm{h}(\yb)  \,. \label{eqn:output_z}
\end{align}

\subsection{Inverse Dynamics}

The framework of servo-constraints is applied to compute the inverse model~\cite{BlajerKolodziejczyk04,Druecker22}. For this purpose, the equations of motion~(\ref{eqn:ode_eqm1})\textendash(\ref{eqn:ode_eqm3}) are appended by the servo-constraints
\begin{align}
\bm{s}(\yb,t) = \bm{h}(\yb)-\zb_{\rm d}(t) &= \bm{0} \label{eqn:sc} \,,
\end{align}
which enforce the system output~$\zb$ to equal the sufficiently smooth desired trajectory~$\zb_{\rm d}$. The resulting differential-algebraic equations 
\begin{align}
\ybp &= \Zb(\yb) \vb \label{eqn:scdae1} \\
\Mb(\yb,t) \, \vbp + \kb(\yb,\vb,t) &= \qb(\yb,\vb,t) + \Cb(\yb,\vb,t)^{\rm T} \lamb + \Bb(\yb) \, \ub  \label{eqn:scdae2} \\
\cb(\yb,\vb,t) &= \bm{0} \,,  \label{eqn:scdae3} \\
\bm{s}(\yb,t) = \bm{h}(\yb)-\zb_{\rm d}(t) &= \bm{0}  \,  \label{eqn:scdae4} 
\end{align}
describe the inverse model. The solution of Eqs.~(\ref{eqn:scdae1})\textendash(\ref{eqn:scdae4}) includes the control input~$\bm{u}$, which can be directly used as feedforward control $\uffw$. For minimum phase systems (i.e. stable internal dynamics), the DAEs~(\ref{eqn:scdae1})\textendash(\ref{eqn:scdae4}) can be integrated forward in time by suitable DAE solvers, see e.g. \cite{Hairer02}. For non-minimum phase systems, the internal dynamics would become unbounded and stable inversion must be applied \cite{Sastry99,ChenPaden96}. 

\section{Stable Inversion}\label{sec:3}

For non-minimum phase systems, the inverse model problem~(\ref{eqn:scdae1})\textendash(\ref{eqn:scdae4}) cannot be solved by forward time integration and stable inversion must be applied. This is the case for many flexible manipulators during end-effector tracking. Stable inversion is proposed in \cite{ChenPaden96} to compute a bounded solution for the inverse model problem. It is formulated for the explicitly stated internal dynamics. The approach is extended to inverse models described by servo-constraints in~\cite{BruelsEtAl13}. Experimental results of the concept are shown in \cite{BurkhardtEtAl15,MorlockEtAl22} for a flexible manipulator. In the following, stable inversion is briefly reviewed and an approximation of the original boundary conditions is proposed. A convergence analysis for a manipulator with one passive joint supports the proposed approximation.  

\subsection{Original Formulation}
For the derivation of the original stable inversion problem, the equations of motion~(\ref{eqn:ode_eqm1})\textendash(\ref{eqn:ode_eqm2}) are considered for multibody systems in minimal coordinates without any geometric constraints ($\nc=0$). Then, the equations of motion can be summarized for notational simplicity in the input affine form
\begin{align}
\xbp &= \fb(\xb) + \sum\limits_{i=1}^{m}g_i(\xb) \, u_i  \,, \nonumber \\
z_1 &= h_1(\xb) \,, \label{eqn:generalsystem1} \\
\vdots \nonumber \\
z_m &= h_m(\xb) \nonumber 
\end{align}
with the states~$\xb=\begin{bmatrix} \yb^{\rm T} & \vb^{\rm T} \end{bmatrix}^{\rm T}$ and with $m$ inputs~$u_i$ and $m$ outputs~$z_i$. The relative degree of a system described by equation~(\ref{eqn:generalsystem1}) is the nonlinear extension to the concept of pole excess of linear systems. It is a property of the system dynamics as well as the chosen system output. For the considered multi-input multi-output systems, the vector relative degree $\rb=\{r_1,r_2,\dots,r_m\}$ is considered. It is defined by the number of Lie derivatives taken of each system output $z_i\,\text{with}~ i = 1,2,\dots,m$, until at least one system input appears explicitly. This is described mathematically as
\begin{align}
z_i^{(k)} &= L_{\bm{f}}^k \, h_i (\xb) + \sum\limits_{j=1}^{m} \underbrace{L_{g_j} L_{\bm{f}}^{k-1} h_i(\xb) }_{=0}\, u_j  =  L_{\bm{f}}^k \, h_i (\xb) \,, ~~~ 0\leq k \leq r_i - 1 \\
z_i^{(r_i)} &= L_{\bm{f}}^{r_i} h_i(\xb) + \sum\limits_{j=1}^{m} \underbrace{L_{g_j} L_{\bm{f}}^{r_i-1} h_i(\xb)}_{\neq 0} \, u_j  \,,
\end{align}
for $i=1,\dots,m$. Moreover, the coupling matrix between the input and output channels must be regular, refer to~\cite{Sastry99,SlotineLi91} for details. In order to extract the internal dynamics explicitly, a nonlinear coordinate transformation is performed. For this purpose, the outputs~$z_i$ and their first $r_i$ Lie derivatives are chosen as new coordinates. Since the sum of the entries of the vector relative degree~$r=\sum_{j=1}^{m} r_j$ is not necessarily equal to the number of states $2f$, a number of $2f-r$ coordinates must be chosen such that the coordinate transformation is at least a local diffeomorphism. These additional coordinates are called $\etab\in\mathbb{R}^{2f-r}$ and describe the internal dynamics. Performing the coordinate transformation to the system~(\ref{eqn:generalsystem1}) determines the internal dynamics
\begin{align}
\etabp &= \rhob \left( \zb, \zbp, \dots , \zb^{(\max(r_i))} , \etab \right) \,. \label{eqn:int_dyn}
\end{align}
The internal dynamics is driven by the output trajectory~$\zb$. Stability analysis of the nonlinear internal dynamics is usually difficult and is therefore performed for the zero dynamics, defined by zeroing the output with~$\zb=\bm{0}$ and all its derivatives \cite{Sastry99,SlotineLi91}. Then, the zero dynamics is
\begin{align}
\etabp &= \rhob \left( \bm{0}, \bm{0}, \dots , \bm{0} , \etab \right) \,. \label{eqn:int_dyn_zero}
\end{align}
which can be linearized around the equilibrium~$\etab_{\rm eq}$. It is assumed that the zero dynamics has an hyperbolic equilibrium point with $n^{\rm s}$ eigenvalues with negative real part and $n^{\rm u}$ eigenvalues positive real part. Details for the derivation of the internal dynamics can be found in \cite{Sastry99,SlotineLi91} for general nonlinear systems, in \cite{Seifried14} for multibody systems in ODE form and in \cite{Berger17} for nonlinear DAEs. 

The stable inversion approach is stated originally for internal dynamics in the form~(\ref{eqn:int_dyn}). In order to compute a bounded solution of the internal dynamics, boundary conditions are defined such that the initial state starts on the unstable manifold of the equilibrium point and the final state reaches the equilibrium on the stable manifold. Thereby, the stable and unstable manifolds are locally approximated by the stable and unstable eigenspaces of the linearized zero dynamics. The boundary conditions are then given as
\begin{align}
\Bb_{\rm u}^{\text{ode}} \left( \etab(T_0) - \etab_{\rm eq} \right) &= 0 \label{eqn:stableinv_bcs1}  \\
\Bb_{\rm s}^{\text{ode}} \left( \etab(T_{\rm f}) - \etab_{\rm eq}\right)&= 0\,.  \label{eqn:stableinv_bcs2}
\end{align}
Thereby,~$T_0$ and $T_{\rm f}$ denote the initial and final simulation time and $\etab_{\rm eq}$ is the equilibrium. The matrices $\Bb_{\rm s}^{\text{ode}}\in\mathbb{R}^{n^{\rm s}\times (2f-r)}$ and $\Bb_{\rm u}^{\text{ode}}\in\mathbb{R}^{n^{\rm u}\times (2f-r)}$ contain the eigenvectors associated with the $n^{\rm s}$ stable and $n^{\rm u}$ unstable eigenvalues of the zero dynamics respectively, see e.g. \cite{BruelsEtAl13} for a detailed derivation of the respective matrices. The solution of the boundary value problem is non-causal, in the sense that the control input~$\uffw$ induces motion before the start of the trajectory at time $t_0$, which is called pre-actuation. Moreover, a post-actuation is necessary in order to bring the internal dynamics to rest after the end of the trajectory at time $t_{\rm f}$. Therefore, the simulation time interval $\left[T_0,\,T_{\rm f}\right]$ is chosen larger than the interval $\left[t_0,\,t_{\rm f}\right]$ of the desired trajectory. It holds~$T_0\leq t_0$ and $T_{\rm f}\geq t_{\rm f}$.

A similar boundary value problem can be formulated directly for the inverse model~(\ref{eqn:scdae1})\textendash(\ref{eqn:scdae4}) described by servo-constraints~\cite{BruelsEtAl13}. The boundary conditions~(\ref{eqn:stableinv_bcs1})\textendash(\ref{eqn:stableinv_bcs2}) are then chosen accordingly for the complete vector of unknowns of the inverse model problem as 
\begin{align}
\Bb^{\text{dae}}_{\rm u} \left( \xb(T_0) - \xb_{\rm eq} \right) &= 0  \label{eqn:stableinv_bcs_dae1}  \\
\Bb^{\text{dae}}_{\rm s} \left( \xb(T_{\rm f}) - \xb_{\rm eq}\right)&= 0 \,.
\label{eqn:stableinv_bcs_dae2}
\end{align}
with the matrices $\Bb_{\rm s}^{\text{dae}}\in\mathbb{R}^{n^{\rm s}\times (2n+\nc+m)}$ and $\Bb_{\rm u}^{\text{dae}}\in\mathbb{R}^{n^{\rm u}\times (2n+\nc+m)}$ 
and with $\xb$ collecting all unknown variables~$\xb = \begin{bmatrix}\yb^\trans&\vb^\trans&\lamb^\trans&\ub^\trans\end{bmatrix}^\trans$ of the inverse model DAEs~(\ref{eqn:scdae1})\textendash(\ref{eqn:scdae4}). 

The derivation of the boundary conditions~(\ref{eqn:stableinv_bcs1})\textendash(\ref{eqn:stableinv_bcs2}) or ~(\ref{eqn:stableinv_bcs_dae1})\textendash(\ref{eqn:stableinv_bcs_dae2}) is not straightforward for general nonlinear multibody systems, since they depend on the internal dynamics. This limits the application of the approach to systems with few degrees of freedom, for which the internal dynamics can be derived explicitly.

\subsection{Approximation of Boundary Conditions}

An approximation of the boundary conditions~(\ref{eqn:stableinv_bcs1})\textendash(\ref{eqn:stableinv_bcs2}) or (\ref{eqn:stableinv_bcs_dae1})\textendash(\ref{eqn:stableinv_bcs_dae1}) is proposed in the following. Instead of enforcing the state vector to lie on the stable and unstable manifolds, part of the state vector is directly constrained onto the equilibrium point. For the explicitly given internal dynamics~(\ref{eqn:int_dyn}) the approximating boundary conditions are of the form
\begin{align}
\Lb_0^{\text{ode}} \, \etab\left(T_0\right) &=\Lb_0^{\text{ode}} \, \etab_{\rm eq}  \label{eqn:simpl_bcs1} \\
\Lb_{\rm f}^{\text{ode}} \, \etab\left(T_{\rm f}\right) &= \Lb_{\rm f}^{\text{ode}} \,  \etab_{\rm eq}   \, \label{eqn:simpl_bcs2}
\end{align}
with the binary matrices~$\Lb_0^{\text{ode}}\in\mathbb{R}^{n^{\rm a} \times (2f-r)}$ and $\Lb_{\rm f}^{\text{ode}}\in\mathbb{R}^{n^{\rm b} \times (2f-r)}$ selecting $n^{\rm a}$ states to be equal to the equilibrium~$\etab_{\rm eq}$ at time $T_0$ and $n^{\rm b}$ states to be equal to the equilibrium at time $T_{\rm f}$. Thereby it is $n^{\rm b},n^{\rm a}>0$. In total, a number of $n^{\rm a}+n^{\rm b}=2f-r$ conditions are given, which equals the number of unknowns.

Equivalently for the inverse model described by the servo-constraints DAEs~(\ref{eqn:scdae1})\textendash(\ref{eqn:scdae4}), the approximating boundary conditions are proposed as 
\begin{align}
\Lb_{0}^{\text{dae}} \, \xb\left(T_0\right) &=\Lb_{0}^{\text{dae}} \, \xb_{\rm eq}  \label{eqn:simpl_bcs_dae1} \\
\Lb_{\rm f}^{\text{dae}} \, \xb\left(T_{\rm f}\right) &= \Lb_{\rm f}^{\text{dae}} \,  \xb_{\rm eq}   \,
\label{eqn:simpl_bcs_dae2}
\end{align}
with the binary matrices~$\Lb_0^{\text{dae}}\in\mathbb{R}^{n^{\rm a} \times (2n+\nc+m)}$ and $\Lb_{\rm f}^{\text{dae}}\in\mathbb{R}^{n^{\rm b} \times (2n+\nc+m)}$ again selecting $n^{\rm a}>0$ states to be equal to the equilibrium~$\etab_{\rm eq}$ at time $T_0$ and $n^{\rm b}>0$ states to be equal to the equilibrium at time $T_{\rm f}$. The number of boundary conditions again equals the number of unknowns with $n^{\rm a}+n^{\rm b}=2n+\nc+m$.

The proposed simplified boundary conditions~(\ref{eqn:simpl_bcs1})\textendash(\ref{eqn:simpl_bcs2}) and (\ref{eqn:simpl_bcs_dae1})\textendash(\ref{eqn:simpl_bcs_dae2}) approximate the correct boundary conditions for small values of the state vectors $\etab$ and $\xb$ respectively. The approximation reduces the effort for system analysis and makes stable inversion applicable for complex multibody systems.

\subsection{Convergence of Approximated BVP}

\begin{figure}[tb]
\centering
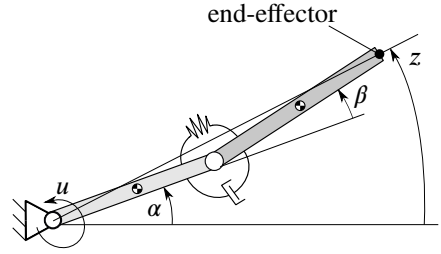
\caption{Model of a manipulator with one passive joint.}
\label{fig:TwoArm_model}
\end{figure}

Convergence of the proposed approximation is here shown graphically for a robot with one passive joint. The robot model is shown in Fig.~\ref{fig:TwoArm_model}. It consists of two links, which are connected by a linear spring-damper combination. The minimal coordinates are chosen as $\alp$ describing the angle of the first link and $\beta$ describing the angle of the second link relative to the first link. The system input is a torque applied on the first joint, while the system output~$z$ is the angle between end-effector and the horizontal line. The simulation parameters are given in Tab.~\ref{tab:TwoArm_SimParam}. The internal dynamics of the system can be described by the coordinate~$\beta$ of the passive joint and it can be derived analytically, see~\cite{Seifried12} for detailed derivations. Here, the analytical solution for the internal dynamics is used to compute a reference solution of the stable inversion approach based on the original boundary conditions. The internal dynamics of the robot is unstable for a homogeneous mass distribution and the output considered here. Therefore, stable inversion with the original boundary conditions as well as with the approximated boundary conditions is applied to compute the feedforward control input in the following. The boundary value problems are solved using finite differences with Simpson discretization. 

\begin{table}[tb]
\caption{Simulation parameters of the manipulator with one passive joint.}
\centering
\label{tab:TwoArm_SimParam}
    \begin{tabularx}{4.6cm}{cc}
	\toprule
    Parameter &  Value  \\
	\midrule
	$L_1 = L_2$ & \SI{0.5}{\metre} \\
	$m_1 = m_2$ & \SI{0.05}{\kilogram} \\
	$d$ & \SI{2.5e-5}{\newton\metre\second\per\radian} \\
	$k$ & \SI{0.5}{\newton\metre\per\radian} \\
	\bottomrule
    \end{tabularx}
\end{table}

\begin{figure}[htb]
\centering
\subfloat[][Desired trajectory and simulated system output~$z$.]{
\includegraphics{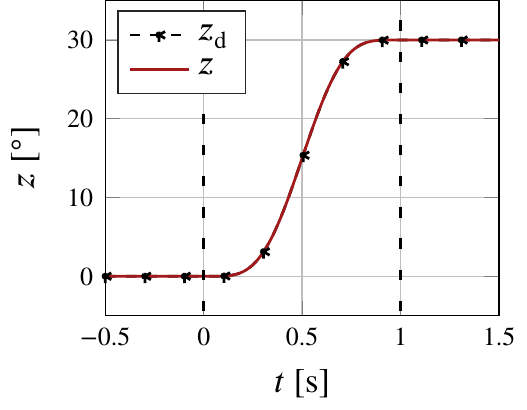}
\label{fig:TwoArm_simoutput}}
\subfloat[][Desired system input~$u_{\rm ffw}$.]{
\includegraphics{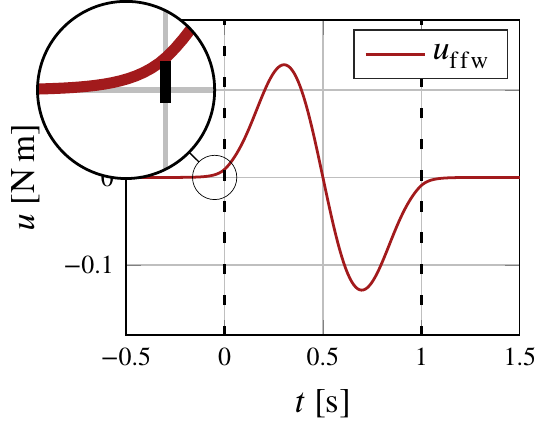}
\label{fig:TwoArm_siminput}}
\subfloat[][Phase space of internal dynamics.]{
\includegraphics{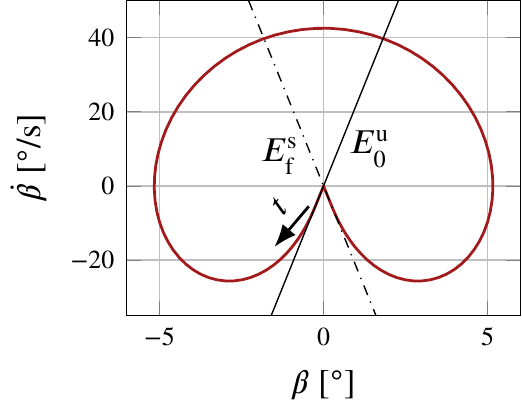}
\label{fig:TwoArm_phasespace}}
\caption{Stable inversion results for the manipulator with one passive joint.}
\label{fig:TwoArm_simres}
\end{figure}

The desired output trajectory $z_{\rm d}$ is chosen as smooth transition from~$z(t_0)=0\,^\circ$ to~$z(t_{\rm f})=30\,^\circ$. The initial and final time are chosen as~$t_0=\SI{0}{\second}$ and $t_{\rm f}=\SI{1}{\second}$. The desired trajectory is shown in Fig.~\subref*{fig:TwoArm_simoutput}. The system input computed by the original stable inversion approach is shown in Fig.~\subref*{fig:TwoArm_siminput}. It can be seen that there exists a pre-actuation phase before $t_0$. This is necessary to obtain a bounded solution for the internal dynamics. The phase space is shown in Fig.~\subref*{fig:TwoArm_phasespace}. It can be seen that the state trajectory leaves the equilibrium in the direction of the unstable eigenspace, denoted by $E_0^{\rm u}$ and reaches the equilibrium in the direction of the stable eigenspace, denoted by $E_{\rm f}^{\rm s}$, as it is enforced by the boundary conditions. 

In the following, the approximating BVP is compared to the original BVP. The approximated solution is denoted by \textit{approx}, while the original solution is denoted by \textit{orig}. For the approximated boundary conditions, the matrices of Eqs.~(\ref{eqn:simpl_bcs1})\textendash(\ref{eqn:simpl_bcs2}) are chosen as
\begin{align}
\Lb_0^{\text{ode}} &= \begin{bmatrix} 1 & 0 \end{bmatrix} \,, \\
\Lb_{\rm f}^{\text{ode}} &= \begin{bmatrix} 1 & 0  \end{bmatrix}   \,.
\end{align}
Therefore, the angle $\beta$ is fixed to the equilibrium point both at the beginning and the end of the trajectory and $\dot{\beta}$ is free. Note that the other possible boundary conditions, e.g. $\Lb_0^{\text{ode}} = \begin{bmatrix} 0 & 1 \end{bmatrix}$ and $\Lb_{\rm f}^{\text{ode}} = \begin{bmatrix} 1 & 0  \end{bmatrix}$, yield similar solutions.

\begin{figure}[tb]
\centering
\subfloat[][System input~$u$.]{
\includegraphics{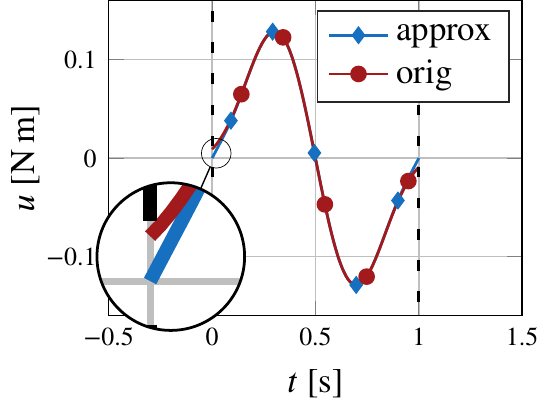}
\label{fig:twoarm_input_changet0tf}}
\subfloat[][Phase space of internal dynamics.]{
\includegraphics{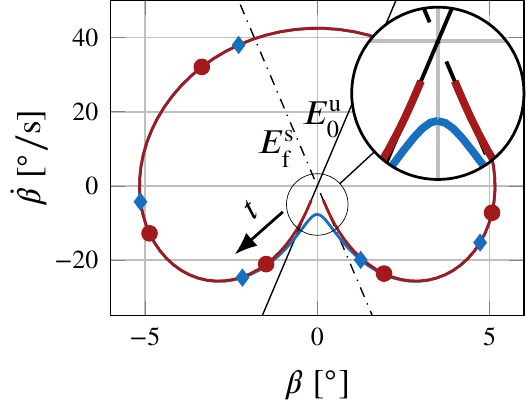}
\label{fig:twoarm_phasespace_changet0tf}}
\caption{Simulation results for the interval $T_0=t_0$ and $T_{\rm f}=t_{\rm f}$.}
\label{fig:TwoArm_changet0tf}
\end{figure}

Figure \ref{fig:TwoArm_changet0tf} shows the results for the system input and the phase space for a simulation interval  $T_0=t_0$ and $T_{\rm f}=t_{\rm f}$, i.e. no allowed pre- and postactuation phase. It can be seen that the approximated solution is similar to the original solution over the major part of the trajectory. There are some differences in the beginning of the trajectory due to the approximated boundary condition. Choosing part of the state vector to equal the equilibrium point introduces the error, because the state trajectory cannot start on the unstable manifold. This can be seen in the phase space diagram in Fig.~\subref*{fig:twoarm_phasespace_changet0tf} in the area around the equilibrium~$\beta=0$. 

\begin{figure}[htb]
\centering
\subfloat[][System input~$u$.]{
\includegraphics{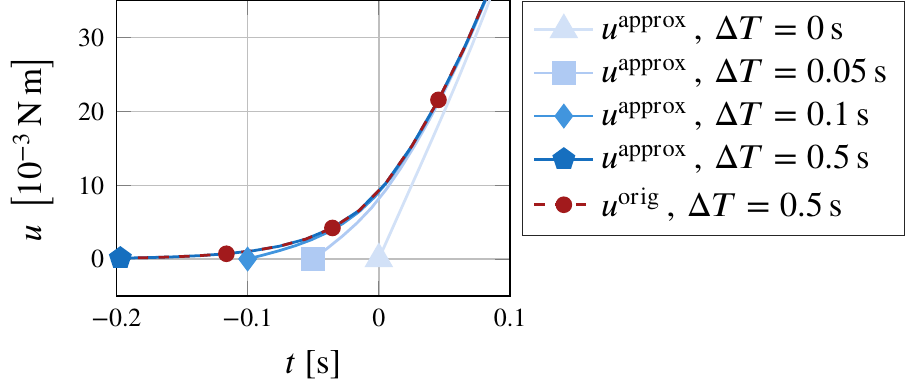}
\label{fig:TwoArm_input_changet0tf2}}
\subfloat[][Error $e$.]{
\includegraphics{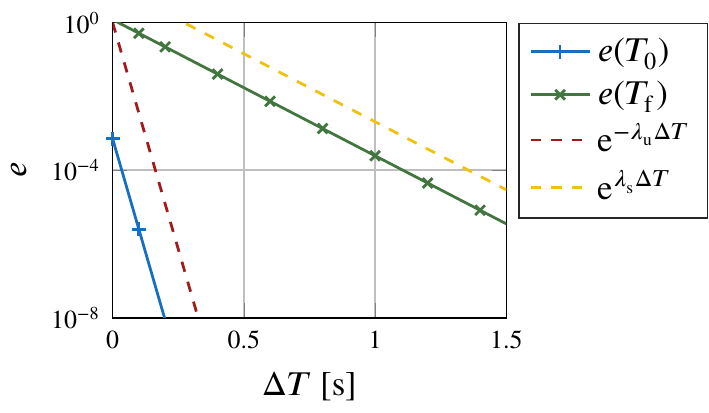}
\label{fig:twoarm_convergence2}}
\caption{Convergence of the approximated solution to the correct solution for increasing simulation interval~$\Delta T$.}
\label{fig:convergence}
\end{figure}

Increasing the simulation interval, such that $T_0<<t_0$ and $T_{\rm f}>>t_{\rm f}$ shows the convergence of the solution. Thereby, the solution interval $\left[T_0,\,T_{\rm f}\right]$ is increased symmetrical around the interval $\left[t_0,\,t_{\rm f}\right]$ with $\Delta T = t_0-T_0 = T_{\rm f}-t_{\rm f}$. The convergence of the system input for larger $\Delta T$ is shown in Fig.~\subref*{fig:TwoArm_input_changet0tf2} for the input trajectory. For this example, a pre- and postactuation phase of $\Delta T = \SI{0.5}{\second}$ is sufficient to recover the solution based on the original boundary conditions. The convergence is shown in terms of the error 
\begin{align}
e(t) = \left\| \etab^{\rm approx}(t) -  \etab^{\rm orig}(t) \right\|
\end{align}
with $\etab = \begin{bmatrix} \beta & \dot{\beta} \end{bmatrix}^{\rm T}$ in Fig.~\subref*{fig:twoarm_convergence2}. The logarithmic convergence diagram shows that the error $e$ converges with the speed of the eigenvalue $\lambda_{\rm u}$ of the internal dynamics at the beginning of the trajectory and with the speed of the eigenvalue $\lambda_{\rm s}$ at the end of the trajectory. Thereby, $\lambda_{\rm u}$ and $\lambda_{\rm s}$ have positive and negative real part respectively, denoting the unstable and stable contributions. Mathematical derivation of the convergence and convergence speed is shown in~\cite{Druecker22}. 

\section{Application Example}\label{sec:4}

For trajectory tracking of highly flexible manipulators, it might not be sufficient to consider only one or two flexible degrees of freedom. Therefore, more complex models should be considered during model inversion. The proposed boundary conditions simplify the application of stable inversion allowing to treat more complex models. As an application example, a flexible manipulator is modeled using the ANCF. The complexity of the equations of motion make an analytical derivation of the internal dynamics burdensome and the original stable inversion formulation is not applicable. The proposed approximation enables the application of stable inversion to the ANCF beams. In the following, the ANCF model is first introduced. Afterwards, simulation results for the approximated stable inversion problem are shown.

\subsection{Flexible Manipulator Modeled by the ANCF} \label{sec:ancf_model}

\begin{figure}[b]
\begin{center}
\includegraphics{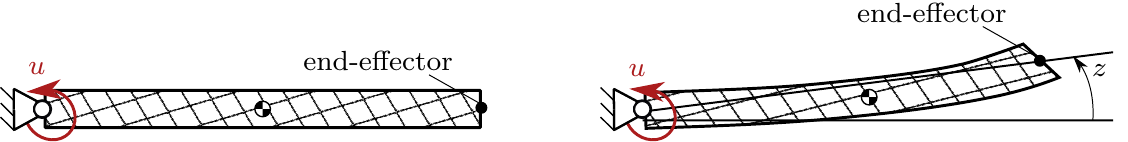}
\caption{Model of a flexible manipulator modeled by the ANCF.}
\label{fig:ancffarb2}
\end{center}
\end{figure}

\begin{figure}[b]
\begin{center}
\includegraphics{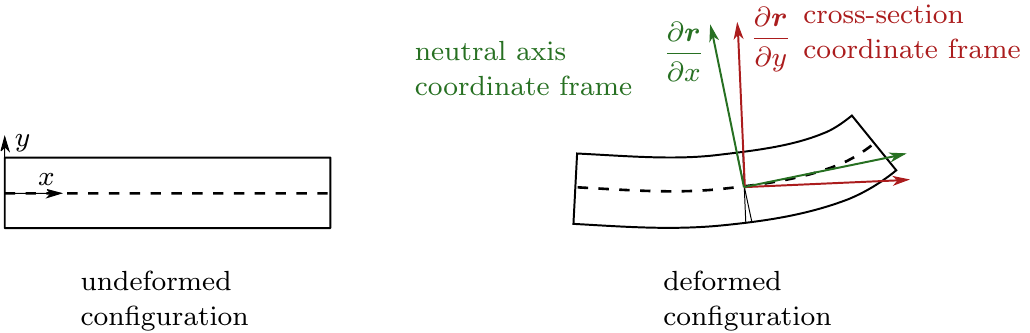}
\caption{Deformed and undeformed configuration of an ANCF element.}
\label{fig:ancf_deformed}
\end{center}
\end{figure}

A flexible manipulator is presented in the following and shown in Fig.~\ref{fig:ancffarb2}. Equivalently to the manipulator discussed above, a single input $u$ is mounted on the left hand joint and actuates the flexible manipulator. The system output $z$ is the angle between end-effector and the horizontal line. 
The flexible manipulator is modeled using the absolute nodal coordinate formulation, which is a nonlinear finite element approach. In contrast to classical finite elements, the ANCF accounts for large rigid body rotations \cite{Shabana13}. Here, the two-dimensional beam element introduced in \cite{OmarShabana01} is considered. The beam element relaxes Euler-Bernoulli assumptions in the sense that shear deformations are allowed and the beam cross-section does not stay perpendicular to the neutral axis \cite{OmarShabana01}. The undeformed and deformed beam configurations are shown in Fig.~\ref{fig:ancf_deformed}. Thereby, the cross-section coordinate frame describes the orientation of the beam cross-section. The neutral axis coordinate frame describes the orientation of the tangent of the neutral axis. 

Following \cite{OmarShabana01}, the ANCF beam is briefly introduced. An arbitrary point on a beam element is described by the position vector
\begin{align}
\rb = \begin{bmatrix}
r_1 \\ r_2
\end{bmatrix} = \Sb(x,y) \, \eb
\end{align}
with the global element shape function $\Sb\in\mathbb{R}^{2\times12}$ and the vector of generalized coordinates~$\yb = \eb\in\mathbb{R}^{12}$. For one ANCF element of length $L$, the generalized coordinates~$e_i$ with $i=1,2,\dots,6$ at the left node are expressed as
\begin{alignat}{4}
e_1 & = r_1 \bigg\rvert_{x=0} \,,\;\;
&e_2 & = r_2 \bigg\rvert_{x=0} \,, && &&  \\
e_3 &= \dfrac{\partial r_1}{\partial x} \bigg\rvert_{x=0} \,,\;\;
&e_4 & = \dfrac{\partial r_2}{\partial x}\bigg\rvert_{x=0} \,,\;\;
e_5 && = \dfrac{\partial r_1}{\partial y}\bigg\rvert_{x=0} \,,\;\;
e_6 && = \dfrac{\partial r_2}{\partial y}\bigg\rvert_{x=0}
\end{alignat}
and the coordinates~$e_i$ with $i=7,8,\dots,12$ at the right node are 
\begin{alignat}{4}
e_7 &= r_1 \bigg\rvert_{x=L} \,,\;\;
&e_8 &= r_2 \bigg\rvert_{x=L}\,, \\
e_9 &= \dfrac{\partial r_1}{\partial x} \bigg\rvert_{x=L} \,,\;\;
&e_{10} & = \dfrac{\partial r_2}{\partial x}\bigg\rvert_{x=L} \,,\;\;
e_{11} && = \dfrac{\partial r_1}{\partial y}\bigg\rvert_{x=L} \,,\;\;
e_{12} && = \dfrac{\partial r_2}{\partial y}\bigg\rvert_{x=L} \,.
\end{alignat}
Thereby, $x$ denotes the undeformed coordinate axis in beam direction with $x\in[0;L]$ and $y$ describes the direction perpendicular to $x$ in the undeformed configuration. For interpolation of the motion, the global shape function matrix is defined as
\begin{align}
\setcounter{MaxMatrixCols}{20}
\Sb(x,y) = \begin{bmatrix}
s_1(x,y) & 0 & s_2(x,y) & 0 & s_3(x,y) & 0 & s_4(x,y) & 0 & s_5(x,y) & 0 & s_6(x,y) & 0 \\
0 & s_1(x,y) & 0 & s_2(x,y) & 0 & s_3(x,y) & 0 & s_4(x,y) & 0 & s_5(x,y) & 0 & s_6(x,y) 
\end{bmatrix} \,.
\end{align}
Thereby, the shape functions are 
\begin{alignat}{3}
s_1(x,y) &= 1-3\xi^2+2\xi^3 \,, \quad
&&s_2(x,y) = L\left(\xi-2\xi^2+\xi^3\right) \,,\quad
&&s_3(x,y) = L\left(\eta - \xi \, \eta\right) \,,\\
s_4(x,y) &= 3\xi^2 - 2\xi^3\,, \quad
&&s_5(x,y) = L\left(-\xi^2 + L\xi^3 \right)\,,\quad
&&s_6(x,y) = L\xi\,\eta \,,
\end{alignat}%
with $\xi = x/L$ and $\eta = y/L$. The matrices of the equations of motion~(\ref{eqn:ode_eqm2}) for one element arise as 
\begin{align}
\Mb &= \int_{V} \rho \, \Sb(x,y)^{\rm T}\,\Sb(x,y) \, \rm{d}V \,, \\
\kb &= \bm{0}  \,, \\
\qb(\yb) &= -\Kb(\eb) \, \eb \label{eqn:strainenergy}
\end{align}
were $\rho$ is the density of the beam, $V$ denotes the volume of one element and $\Kb\in\mathbb{R}^{12\times12}$ is a state dependent stiffness matrix. Note that the mass matrix~$\Mb$ is constant over time and is evaluated before the simulation for computational efficiency. The derivation of the elastic forces~$\qb$ is taken from \cite{Garcia-VallejoEtAl04}, who take a general continuum mechanics approach to model the elastic forces. They arise from the strain energy~$U_e$ as
\begin{align}
\qb(\yb,\vb,t) &= \dfrac{\partial U_e}{\partial \eb} \,.
\end{align}
Substituting a linear material model for the strain energy~$U_e$ and using simplifications shown in~\cite{Garcia-VallejoEtAl04} yields the simplified elastic forces in the form of Eq.~(\ref{eqn:strainenergy}). Note that using this approach, expensive evaluations of the volume integrals at each time step are avoided. Instead, a few invariant matrices are calculated beforehand to simplify the evaluation of the stiffness matrix~$\Kb$. Another advantage of this approach lies in the simple derivation of the Jacobian matrix of the elastic forces, which can be derived completely analytically, see \cite{Garcia-VallejoEtAl04} for details. The analytical Jacobian matrices are also relevant for the efficient numerical solution of the large but sparse boundary value problem.

The joint at the left node, see Fig.~\ref{fig:ancffarb2}, is enforced by the algebraic constraint
\begin{align}
\cb_1(\eb) &= \begin{bmatrix} e_1 \\ e_2  \end{bmatrix} = \bm{0} \,,
\end{align}
and the equations of motion arise in DAE form.
The system input~$u$ is considered as a velocity-controlled actuator acting on the left node. A zero-order hold model is assumed for the actuator, such that the actuator velocity $u$ equals the rotational velocity~$\dot{\gamma}$ of the cross-section at the left joint. Mathematically, this is described by the constraint
\begin{align}
c_2(\yb) =  u - \dot{\gamma} = u + \dfrac{e_6 \,\dot{e}_5 - e_5 \, \dot{e}_6}{f_{56}^2} =0 \,. \label{eqn:ANCF_pt0actuator}
\end{align}
The system output~$z$ is defined as the angle between the right hand node and the horizontal, see Fig.~\ref{fig:ancffarb2} with
\begin{align}
z = \arctan\left(\dfrac{e_{6(N+1)-4}}{e_{6(N+1)-5}}\right)  \,, \label{eqn:chap2:ancf_output}
\end{align}
where $N$ is the number of ANCF beam elements. Note that the considered system output~$z$ is not a function of the actuated coordinates. This property and the complex dynamic equations make a direct derivation of the internal dynamics burdensome.

\subsection{Stable Inversion Results} \label{sec:chap4_ancf}

In the following, stable inversion is applied to the flexible manipulator from Fig.\ref{fig:ancffarb2}. The model parameters for the ANCF beam with squared cross-section are given in Tab.~\ref{tab:ANCFParam}. The boundary value problem is setup for the complete inverse model described by servo-constraints considering the approximating boundary conditions~(\ref{eqn:simpl_bcs_dae1})\textendash(\ref{eqn:simpl_bcs_dae2}). The Simpson scheme is again used for discretization and the step size is chosen as~$h=\SI{0.01}{\second}$. The initial guess for the boundary value problem is obtained by computing the system input $u_{\rm rigid}$ for an equivalent rigid system and applying it to the flexible system with increased stiffness $E=\SI{1.2e9}{\pascal}$ in a forward simulation. Alternatively, the initial guess can for example be obtained by redefining the system output to obtain stable internal dynamics and computing the respective inverse model by forward time integration.

\begin{table}[htb]
\caption{Overview of simulation parameters for the flexible manipulator.}
\label{tab:ANCFParam}
\begin{center}
    \begin{tabularx}{10.5cm}{cccc}
	\toprule
    Material parameter &  Value  & Geometry parameter &  Value  \\
	\midrule
	$\rho$ & \SI{910}{\kilogram\per\metre\cubed} & $L$ & \SI{1}{\metre} \\
	$\nu$ & {0} & $A$ & \SI{0.0081}{\metre\squared} \\
	$E $ & \SI{1.2e7}{\pascal}& & \\
	\bottomrule
    \end{tabularx}
\end{center}
\end{table}

For model inversion, a number of 4 ANCF beam elements are considered in order to limit the computational effort. They yields a number of 27 unactuated coordinates. The system input obtained from model inversion is then applied to a manipulator modeled by 10 ANCF elements in a forward simulation, since convergence of the ANCF model is reached for approximately 10 elements.

\begin{figure}[htbp]
\centering
{\subfloat[Feedforward input~$u_{\rm ffw}$.\label{fig:ANCFtwoarm_input_Elow}]%
{\includegraphics{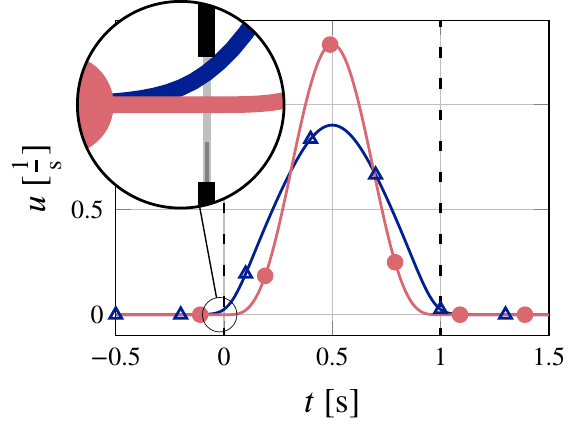}}}
{\subfloat[Simulated system output $z$. \label{fig:ANCFtwoarm_output_Elow}]%
{\includegraphics{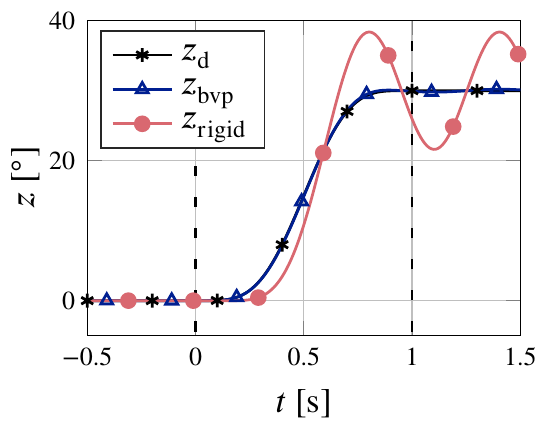}}}
{\subfloat[Tracking error $e=z-z_{\rm d}$. \label{fig:ANCFtwoarm_outputerror_Elow}]%
  {\includegraphics{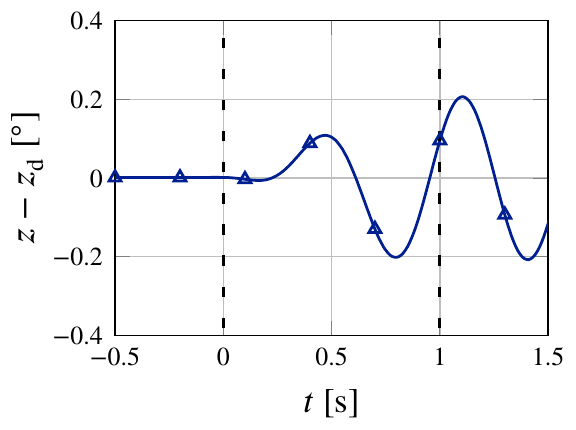}}}
\caption{Simulation results  for one ANCF beam element in configuration~2.}
\label{fig:ancf_elow}
\end{figure}

\begin{figure}[htbp]
\centering
\includegraphics{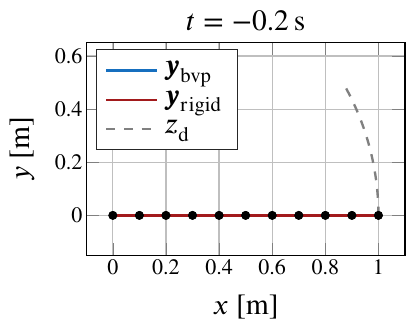}
\includegraphics{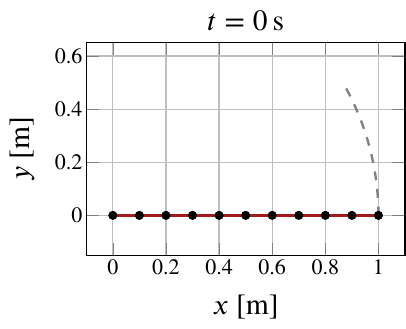}
\includegraphics{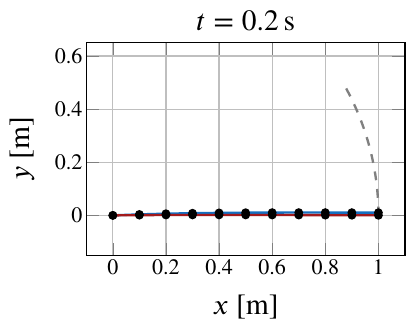}\\
\includegraphics{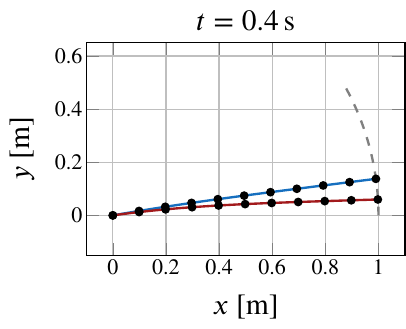}
\includegraphics{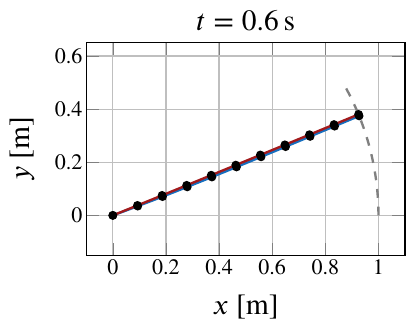}
\includegraphics{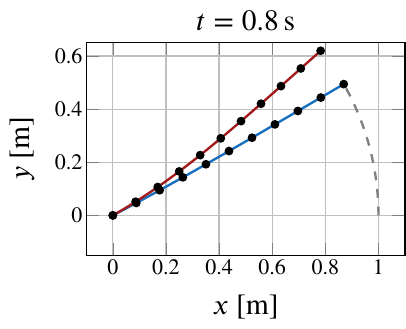}\\
\includegraphics{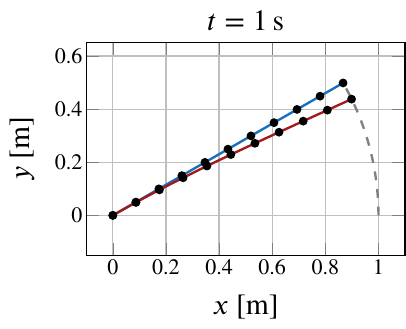}
\includegraphics{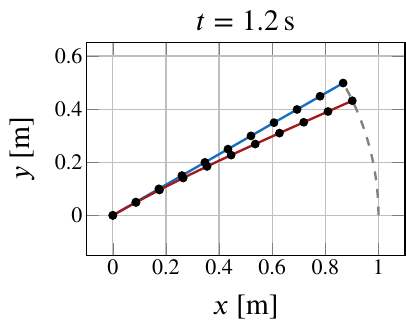}
\includegraphics{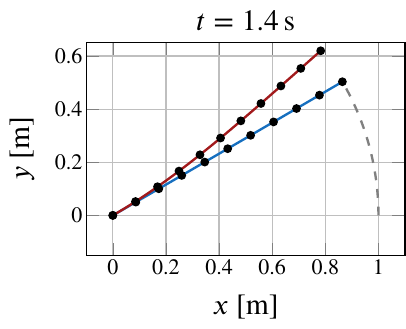}
\caption{Visualization of the motion in space for different time instances.\label{fig:ANCF_xyspace}}
\end{figure}

The desired trajectory is chosen as before, see Fig.~\subref*{fig:TwoArm_simoutput}. The  results of the stable inversion are shown in Fig.~\ref{fig:ancf_elow}. They are compared to a simulation with the system input~$u_{\rm rigid}$, which is obtained from inverting an equivalent rigid beam. The system input~$u_{\rm ffw}$ obtained from stable inversion features a preactuation phase and differs from the input for the equivalent rigid system, see Fig.~\subref*{fig:ANCFtwoarm_input_Elow}. This is also reflected in the simulated system output~$z$ in Fig.~\subref*{fig:ANCFtwoarm_output_Elow}. The system input~$u_{\rm rigid}$ induces oscillations around the desired final position of approximately $8^\circ$, because the flexibility is not accounted for during model inversion. In contrast, the system input from stable inversion of the flexible system captures the dynamics well and yields accurate tracking results in forward simulation. Even though the inversion was performed for only 4 ANCF beam elements, the forward simulation with 10 beam elements shows very accurate tracking. The tracking error $e=z-z_{\rm d}$ is shown in Fig.~\subref*{fig:ANCFtwoarm_outputerror_Elow} and has a maximum value of $e=0.2^\circ$. First of all, the results show that the model inversion with a low number of ANCF elements is reasonable to save computational effort, since the tracking is sufficiently accurate. In contrast, the results demonstrate that inverting the rigid beam is not sufficient for accurate trajectory tracking. In order to visualize the results, the simulation results are shown in $x,y$-space for different time instances in Fig.~\ref{fig:ANCF_xyspace}. 

\section{Summary}
\label{sec:5}

Flexible manipulators arise naturally during the design of lightweight machines. These systems are often underactuated and the trajectory control of such systems is an active field of research. The control is  difficult due to the non-minimum phase behavior of many common flexible manipulators. This makes forward integration of the inverse model impossible. Alternatively, the stable inversion approach can be applied. However, the original formulation depends on deriving the internal dynamics explicitly. This is not straightforward for complex multibody systems, such as very flexible manipulators. Therefore, an approximation of the boundary value problem is proposed, which does not rely on deriving the internal dynamics explicitly. It is demonstrated that the approximation converges to the original solution for a simple system with one passive joint. The approximation is then applied to solve the inverse model problem of a highly flexible manipulator, for which it is not possible to determine the internal dynamics. As an application example, the flexible manipulator is modeled using the ANCF. The numerical results show that the proposed approximation yields accurate results for the ANCF model. The results show that superior tracking can be achieved when considering the flexible system during model inversion compared to simply inverting the equivalent rigid system.

\section*{Funding}
This work was supported by the German Research Foundation (Deutsche Forschungsgemeinschaft) via the grant 362536361.

\section*{Conflict of Interest}
 The authors declare that they have no conflict of interest.

\bibliography{literature}%

\end{document}

%% file: colors.tex
\definecolor{mypurple}{rgb}{0.400,0.00,0.400}

\definecolor{myblue}{rgb}{0.086,0.435,0.749}
\definecolor{mymediumblue}{rgb}{0.2549,0.5843,0.8706}
\definecolor{mylightblue}{rgb}{0.686,0.792,0.953}
\definecolor{myverylightblue}{rgb}{0.8235,0.8824,0.9686}

\definecolor{myred}{rgb}{0.6353    0.1020    0.1098}
\definecolor{mylightred}{rgb}{0.85,0.4078,0.439}

 
\definecolor{mylightgreen}{rgb}{0.2823,0.611,0.3294}
\definecolor{mygreen}{rgb}{0.2549, 0.4667, 0.2471}

\definecolor{mygrey}{rgb}{0.5,0.5,0.5}
\definecolor{mycolorgrey}{rgb}{0.5,0.5,0.5}

\definecolor{myyellow}{rgb}{0.9412 ,0.7608, 0.0627}

%% file: TwoArm_z.pdf_tex
\begingroup%
  \makeatletter%
  \providecommand\color[2][]{%
    \errmessage{(Inkscape) Color is used for the text in Inkscape, but the package 'color.sty' is not loaded}%
    \renewcommand\color[2][]{}%
  }%
  \providecommand\transparent[1]{%
    \errmessage{(Inkscape) Transparency is used (non-zero) for the text in Inkscape, but the package 'transparent.sty' is not loaded}%
    \renewcommand\transparent[1]{}%
  }%
  \providecommand\rotatebox[2]{#2}%
  \newcommand*\fsize{\dimexpr\f@size pt\relax}%
  \newcommand*\lineheight[1]{\fontsize{\fsize}{#1\fsize}\selectfont}%
  \ifx\svgwidth\undefined%
    \setlength{\unitlength}{161.8309142bp}%
    \ifx\svgscale\undefined%
      \relax%
    \else%
      \setlength{\unitlength}{\unitlength * \real{\svgscale}}%
    \fi%
  \else%
    \setlength{\unitlength}{\svgwidth}%
  \fi%
  \global\let\svgwidth\undefined%
  \global\let\svgscale\undefined%
  \makeatother%
  \begin{picture}(1,0.56988332)%
    \lineheight{1}%
    \setlength\tabcolsep{0pt}%
    \put(0,0){\includegraphics[width=\unitlength,page=1]{TwoArm_z.pdf}}%
    \put(0.10675127,0.12867856){\color[rgb]{0,0,0}\makebox(0,0)[lt]{\lineheight{0}\smash{\begin{tabular}[t]{l}$u$\end{tabular}}}}%
    \put(0,0){\includegraphics[width=\unitlength,page=2]{TwoArm_z.pdf}}%
    \put(0.91724458,0.4269489){\color[rgb]{0,0,0}\makebox(0,0)[lt]{\lineheight{0}\smash{\begin{tabular}[t]{l}$z$\end{tabular}}}}%
    \put(0,0){\includegraphics[width=\unitlength,page=3]{TwoArm_z.pdf}}%
    \put(0.45566392,0.52601397){\color[rgb]{0,0,0}\makebox(0,0)[lt]{\lineheight{0}\smash{\begin{tabular}[t]{l}end-effector\end{tabular}}}}%
    \put(0,0){\includegraphics[width=\unitlength,page=4]{TwoArm_z.pdf}}%
    \put(0.31622597,0.07854323){\color[rgb]{0,0,0}\makebox(0,0)[lt]{\lineheight{0}\smash{\begin{tabular}[t]{l}$\alpha$\end{tabular}}}}%
    \put(0,0){\includegraphics[width=\unitlength,page=5]{TwoArm_z.pdf}}%
    \put(0.79156841,0.33936936){\color[rgb]{0,0,0}\makebox(0,0)[lt]{\lineheight{0}\smash{\begin{tabular}[t]{l}$\beta$\end{tabular}}}}%
    \put(0,0){\includegraphics[width=\unitlength,page=6]{TwoArm_z.pdf}}%
  \end{picture}%
\endgroup%